\documentclass[10pt,a4paper,twoside]{article}
\usepackage{JISE}

\usepackage{wrapfig}
\usepackage{times}
\usepackage{verbatim}
\usepackage{color}
\usepackage{url}
\usepackage{graphicx}
\usepackage{array}
\usepackage{longtable}
\usepackage{multicol}
\usepackage{multirow}
\usepackage{pdfpages} % include outside .pdf
\usepackage{wallpaper} % watermark
\usepackage{mathtools}
\usepackage{amsmath}
\usepackage{amssymb}
\usepackage{booktabs}
\usepackage{adjustbox}
\usepackage{newtxtext}
\usepackage{caption}
\usepackage{subcaption}
\usepackage{placeins}
\usepackage{amsthm}

\usepackage[linesnumbered,ruled,vlined]{algorithm2e}
\usepackage[dvipsnames]{xcolor} % color
\usepackage{listings} % code
\usepackage{courier} % for Courier font
\newcommand\pythonstyle{\lstset{
    language=Python,
    basicstyle=\small\ttfamily, % Using Courier font
    morekeywords={self,with}, % Add keywords here
    emph={MyClass,__init__}, % Custom highlighting
    emphstyle=\bfseries\color{BrickRed}, % Using \bfseries for custom highlighting style
    frame=single, % Add frame around the code
    showstringspaces=false,
    morecomment=[l][\itshape\color{Gray}]{\#},
    numbers=left,
    stepnumber=1,
}}
\newcommand\pythoninline[1]{%
    {\pythonstyle\lstinline!#1!}%
}
\lstnewenvironment{python}[1][]{%
    \pythonstyle\footnotesize\lstset{#1}}{}
\usepackage{titlesec}
\titlespacing*{\paragraph}{\parindent}{1ex plus 0.5ex minus 0.2ex}{1em}
\titlespacing*{\section}{0pt}{2ex plus 1ex minus 0.5ex}{1em}

\usepackage[colorlinks,linkcolor=blue,citecolor=blue,bookmarks=false,hypertexnames=true]{hyperref}

\usepackage{tikz}
\usepackage{tikz-qtree}
\usepackage{ifthen}
\newcommand{\verbosemode}{0} % set to 0 to hide text
\newcommand{\verbose}[1]{%
    \ifthenelse{\equal{\verbosemode}{1}}{#1}{}%
}
\usepackage{pgfplots}
\pgfplotsset{compat=1.14}

\newtheorem{theorem}{Theorem}[section]

\newtheorem{corollary}[theorem]{Corollary}

\theoremstyle{definition}
\newtheorem{definition}[theorem]{Definition}
\theoremstyle{remark}

\theoremstyle{plain}

\newcommand{\citep}[1]{\cite{#1}}

\title{Concolic Testing on Individual Fairness of Neural Network Models}

\author{Ming-I Huang, Chih-Duo Hong\thanks{\quad Corresponding author}, and Fang Yu\\
  {\small\em National ChengChi University} \\
  {\small\em Taipei, Taiwan}\\
  {\small\em E-mail: \{111356047; chihduo; yuf\}@nccu.edu.tw}\thanks{\quad Chih-Duo Hong is supported by the National Science and Technology Council (NSTC), Taiwan, under grants 112-2222-E004-001-MY3 and 114-2634-F-004-002-MBK. Fang Yu is supported by the NSTC under 113-2221-E-004-010-MY2, 113-2634-F-004-001-MBK, and 114-2634-F-004-002-MBK.}%
}

\authorrunning{Ming-I Huang, Chih-Duo Hong, and Fang Yu}
\titlerunning{Concolic Testing of Neural Network Fairness}

\begin{document}
%\editorfootnote{Received March 11, 2014; revised June 20, 2014; accepted September 28, 2014. \\
%{\small$^+$}Corresponding author}
%Corresponding author should be marked "+" otherwise it will be regarded as the first author.
\maketitle

\begin{abstract}
    We present PyFair, a concolic path exploration framework for detecting counterfactual fairness violations with respect to protected attributes in neural network classifiers. PyFair extends PyCT by constructing a dual network that evaluates two inputs that share nonprotected attributes but differ in protected attributes. This reduces individual discrimination detection to finding an input for which the dual network predicts unequal labels. When the classifier, domain constraints, and decision predicate are exactly encoded in a decidable SMT theory and path exploration is exhausted, the same construction can certify the absence of such witnesses under the encoding. We evaluate PyCT witness search on individual seed inputs and the PyFair dual network construction on benchmark fully connected networks and on models retrained with prior bias mitigation methods. The results show that these techniques find unfairness witnesses missed by baseline testing and detect residual violations in mitigated models, while exhaustive verification remains limited by path explosion and solver cost.
  % \begin{keywords}
  % wireless sensor networks, localization, mobile beacon, mobile anchor, multiple processes
  % \end{keywords}
\end{abstract}

% Your thesis goes here
\section{Introduction}
\label{c:intro}
 Deep Neural Networks (DNNs) are increasingly deployed across diverse applications, from autonomous vehicles to medical diagnostics. While these models often achieve strong performance, their use in high stakes settings raises significant concerns about trustworthiness and fairness. In critical domains such as criminal justice, employment, and financial services, the algorithmic fairness of DNNs has come under intense scrutiny \cite{biswas2021fair, hort2021fairea, zhang2021ignorance}. Notable examples include racial bias observed in the COMPAS recidivism prediction model~\citep{flores2016false} and gender bias evident in Amazon's recruiting model~\citep{dastin2022amazon}. These examples show the importance of addressing fairness issues in neural networks. 
 
 Various forms of discrimination have been acknowledged, including group discrimination~\citep{feldman2015certifying} and individual discrimination~\citep{dwork2012fairness}. Discrimination is typically defined with respect to a set of protected attributes (PAs), such as race and gender, in contrast to non-protected attributes (NPAs). \emph{Group fairness}~\citep{feldman2015certifying} advocates for impartial treatment among protected groups (i.e., the subpopulation defined by a protected attribute) to eliminate group discrimination. While group fairness is relatively easy to measure and has clear policy implications, it can sometimes mask disparities among individuals or lead to reverse discrimination.
 In contrast, \emph{individual fairness}~\citep{dwork2012fairness} dictates that similar individuals should be treated similarly regardless of their membership in protected groups. In the context of machine learning, this means that two inputs differing solely in their PAs should lead to identical model outcomes.
 Extensive efforts have been directed towards enhancing individual fairness at the model level. One prevalent approach is \emph{fairness testing}, which aims to generate efficient test suites before deployment \citep{galhotra2017fairness, saleiro2018aequitas, zhang2020white, zhang2021efficient, Zheng2022NeuronFair}. These discrimination tests can be employed to quantify discrimination or can be utilized for model retraining to alleviate unfairness.
 Despite these efforts, verifying fairness properties in complex DNNs remains challenging due to their nonlinear nature.

This paper presents PyFair, a framework for detecting counterfactual unfairness witnesses with respect to protected attributes in DNN classifiers using the concolic testing tool PyCT~\citep{chen2021pyct,yu2024constraint}. Given a DNN model, PyFair employs concolic execution to generate path constraints related to fairness. By systematically exploring these constraints, PyFair can identify test inputs that are relevant to fairness assessment.
%In our approach, we designate protected attributes as concolic variables based on individual fairness principles. This allows PyCT to systematically generate diverse concrete values by exploring model branches and identifying PA changes that alter classification.
Although PyCT may be used alongside random sampling to evaluate potential discrimination in a DNN, the absence of detected discrimination generally does not guarantee overall model fairness.
%PyCT can only perturb one PA per sample while keeping NPAs fixed. Hence, it only explores test cases whose NPA values coincide with the given sample. 
%Thus, our second key innovation in this work is proposing a dual network architecture to examine any feasible discriminatory instance with identical NPAs but different PAs. 
PyFair strengthens the capabilities of PyCT by constructing a dual network that represents paired inputs with shared non-protected attributes and different protected attributes. When the classifier decision predicate, input domain constraints, and all relevant network operations are faithfully encoded in an SMT (Satisfiability Modulo Theories) theory~\citep{barbosa2022cvc5}, exhaustive exploration can verify the absence of such witnesses under the encoding.\footnote{For example, ReLU networks with linear threshold or argmax decision predicates can often be encoded in linear arithmetic. If a model implementation uses Sigmoid or Softmax, PyFair reasons about the induced classifier decision predicate unless an exact encoding of the numerical operation is explicitly provided.}

We assess PyFair by evaluating network models studied in the literature \citep{biswas2023fairify,zhang2020white,zhang2021efficient}. Our experiments show that PyCT and PyFair identify counterfactual unfairness witnesses with respect to protected attributes in many benchmark models, including cases where Fairify does not report a witness under the reported configuration. We also test models improved by bias mitigation techniques such as ADF \citep{zhang2020white} and EIDIG \citep{zhang2021efficient}, showing that PyFair can still detect residual witnesses in these retrained models. Finally, we evaluate PyFair on artificially constructed fair models, illustrating both the conditional verification workflow and the scalability limits caused by path explosion and solver cost.

This paper makes the following contributions:
\begin{itemize}
    \item We formulate counterfactual unfairness with respect to protected attributes in neural network classifiers as a search problem over paired inputs that share non-protected attributes and differ only in protected attributes.
    \item We introduce a dual network construction that reduces witness search for such counterfactual pairs to finding an input on which the constructed network predicts label disagreement.
    \item We implement this construction using concolic path exploration and SMT solving, and state the conditions under which exhaustive exploration yields a verification result under the encoded semantics.
    \item We evaluate the approach on existing benchmark neural networks and on models retrained by prior bias mitigation methods, showing both its effectiveness in finding witnesses and its current scalability limits.
\end{itemize}

Overall, PyFair detects counterfactual unfairness witnesses with respect to protected attributes and, under exact encoding and exhaustive exploration, can verify their absence within the encoded domain.
%ur framework's dual network architecture, combined with SMT solvers, enables a comprehensive exploration of fairness properties with theoretical guarantees. 
Compared with random sampling or gradient search methods \citep{galhotra2017fairness, saleiro2018aequitas, zhang2020white, zhang2021efficient, Zheng2022NeuronFair}, our approach provides more systematic path exploration but requires greater computational resources and faces scalability challenges when dealing with complex network architectures.

\section{Related Work}
\label{c:related}

\begin{table}[t]
\caption{Summary of related work in fairness testing and verification}
\label{tab:related-work}
\centering
\resizebox{0.85\columnwidth}{!}{
%\resizebox{0.9\columnwidth}{!}{
\begin{tabular}{|l|l|l|}
\hline
\textbf{Approach} & \textbf{Focus} & \textbf{Methodology} \\ 
\hline
THEMIS \citep{galhotra2017fairness} & Fairness testing & Causality-based random sampling \\ 
\hline
AEQUITAS \citep{saleiro2018aequitas} & Fairness testing & Local and global random sampling \\ 
\hline
SymbGen \citep{aggarwal2019black} & Fairness testing & Test case generation using symbolic execution\\ 
\hline
ADF \citep{zhang2020white} & Bias mitigation & Adversarial sampling based on gradient search\\ 
\hline
EIDIG \citep{zhang2021efficient} & Bias mitigation & An ADF variant with momentum optimization \\ 
\hline
NeuronFair \citep{Zheng2022NeuronFair} & Fairness testing & Adversarial sampling via neuron interpretation \\ 
\hline
Fairify \citep{biswas2023fairify} & Fairness verification & Constraint-based witness search \\ 
\hline
PyFair & Testing \& verification & Constraint-guided concolic path exploration \\ 
\hline
\end{tabular}
}
\end{table}

  %Our review of related work encompasses fairness testing, fairness verification, and applications of concolic testing in DNNs.

\emph{Fairness testing.}
    % Fairness Through Unawareness (FTU) suggests removing protected attributes (PAs) during DNN training \citep{dwork2012fairness}. However, ignoring PAs is often ineffective for eliminating discrimination since PAs may correlate with other variables \citep{awwad2020exploring}.
    % Indeed, studies show that a DNN can still identify race with high accuracy even after removing PAs, which indicates that FTU is insufficient for ensuring fairness \citep{li2023faire}. Therefore, accounting for PAs when analyzing and testing model bias is essential.
    Recent research has focused extensively on testing and validating the fairness of DNNs using discriminatory examples.
    %Recent techniques for testing the fairness of DNNs include MAFT, a black-box approach that achieves comparable effectiveness to white-box methods while improving applicability to large-scale networks \citep{wang2024maft}.
    THEMIS \citep{galhotra2017fairness} introduces fairness scores as measurement metrics of fairness and devises a causality-based algorithm for random discriminatory sample generation. While THEMIS uses pure and unguided random sampling, tools like AEQUITAS \citep{saleiro2018aequitas} and SymbGen \citep{aggarwal2019black} offer more targeted generation algorithms to identify fairness violations. AEQUITAS pioneers a two-step approach combining global and local search strategies, while SymbGen exploits symbolic execution and local explanability to generate effective test cases. Adversarial sampling \citep{ige2023adversarial} is also a popular method for analyzing fairness of DNNs.
  ADF \citep{zhang2020white} adopts a two-phase gradient search to identify discriminatory examples. EIDIG \citep{zhang2021efficient} furthermore optimizes ADF by incorporating momentum in the global generation phase and reducing the frequency of gradient calculations in the local generation phase.
  Despite these advancements, ADF and EIDIG suffer from the issues of gradient vanishing and local optima. To address these challenges, NeuronFair \citep{Zheng2022NeuronFair} 
  interprets internal DNN states to guide instance generation and explore decision boundaries.
Unlike black-box methods such as THEMIS and AEQUITAS, which prioritize efficiency through random sampling, PyFair performs concolic path exploration guided by symbolic constraints at a higher computational cost.
Compared to heuristic search methods like ADF and EIDIG, PyFair offers more systematic exploration of discriminatory instances using concolic testing and SMT solvers. Under exact encoding and exhaustive exploration, this search can also yield a conditional verification result. %This leads to the trade-offs between PyFair and other approaches in rigor and scalability.

\emph{Fairness verification.}
  Testing has proven helpful in identifying fairness violations and addressing model deficiencies, but it often falls short of verifying the absence of fairness violations. 
  Most studies in fairness verification have focused on group fairness, as seen in FairSquare \citep{albarghouthi2017fairsquare} and VeriFair \citep{bastani2019probabilistic}.
  John et al.~\cite{john2020verifying} present the first technique for verifying individual fairness of classical machine learning models. For neural networks, LCIFR \citep{ruoss2020learning} certifies individual fairness by formulating it as a local property, which coincides with robustness within a specific distance metric. In contrast, Libra \citep{urban2020perfectly} computes certifications for the global property of causal fairness.
  %Verification poses a significant challenge, particularly with complex neural network models,
  Conceptually, ensuring individual fairness entails verifying a local or global robustness property, wherein the classifier output remains unchanged for perturbations of any input within the domain.
  Fairify \citep{biswas2023fairify} is a constraint-based approach for verifying the individual fairness of DNNs. The tool decomposes the verification task into multiple sub-problems and prunes the networks to mitigate verification complexity.
  While designed primarily for witness search, our method can also verify the absence of discriminatory instances when the classifier semantics and domain constraints are exactly encoded and concolic exploration is exhaustive.
  %The verification process comprises three key steps: 1) Input partitioning, 2) Sound pruning (including two phases for identifying consistently inactive neurons and eliminating them: interval analysis and individual verification), and 3) Heuristic-based pruning.
  %However, a notable drawback of SMT-solvers is their considerable time overhead, especially when handling complex formulas arising from the feed-forward process of DNNs. Consequently, Fairify often surpasses the predefined timeout duration, resulting in an outcome marked as ``UNKNOWN.'' 
  Another line of research attempts to achieve individually fair models through enforcement during model training \citep{yurochkin2019training, ruoss2020learning, li2023faire, khedr2023certifair, mohammadi2023feta}. Although this work focuses primarily on determining whether a pre-trained DNN model violates fairness, our method can be easily integrated into existing fair training approaches. Indeed, discriminatory instances identified by our method can be used in model training and refinement, such as generating challenging test cases to evaluate the trained models and identifying areas where fairness enforcement might fall short.

\emph{Concolic testing.}
  %Addressing the challenges associated with the verification and security of deep neural networks necessitates formal verification techniques such as Reluplex \citep{katz2017reluplex}, Marabou \citep{katz2019marabou}, and ReluVal \citep{wang2018formal}, which can provide guarantees of correctness. Given the computational complexity of these methods, abstraction interpretation techniques like DeepPoly \citep{singh2019abstract} and AI2 \citep{gehr2018ai2} offer promising solutions for ensuring soundness by abstracting neural network model states and computations. However, the scalability of these methods is often limited, highlighting the need to explore alternative approaches, including automatic testing.
  Concolic testing has been adapted for neural networks to explore execution paths and increase test coverage \citep{sun2018concolic}.
  Tools such as DeepXplore \citep{pei2017deepxplore}, DeepGauge \citep{ma2018deepgauge}, and DeepCon \citep{zhou2021deepcon} offer alternative avenues by generating adversarial examples that expose vulnerabilities in neural networks. %The effectiveness of these testing techniques hinges on test inputs that sufficiently cover the behaviors of neural networks. To this end, concolic execution techniques are employed to generate path constraints along a concrete execution systematically. These constraints are then used to find new test inputs, enhancing coverage in the search space.
  DeepXplore introduces neuron coverage as a metric for measuring DNN testing adequacy and uses multiple similar DNNs as cross-referencing oracles to avoid manual checking.
  DeepCon proposes contribution coverage, which considers both neuron outputs and connection weights to gauge testing adequacy.
  DeepConcolic \citep{sun2019deepconcolic} conducts symbolic execution testing based on neuron coverage, generating inputs that activate neurons not triggered in the current execution. By combining gradient-based and constraint-based methods, DeepConcolic systematically maximizes neuron coverage across various paths. %significantly advancing constraint-based testing for neural networks. 
  Most existing DNN testing tools focus on maximizing certain coverage measures instead of exploring critical branches for changing prediction outcomes. As a result, it is not straightforward to plug these tools into our framework to detect discrimination concerning protected attributes. %In our experiments, we utilize PyCT \citep{chen2021pyct} to conduct fairness testing for DNNs.

\section{Discriminatory Instance Checking with PyCT}
\label{c:discriminatory_checking}

\begin{algorithm}[t]
\caption{Concolic witness search for protected-attribute counterfactual unfairness}
\label{alg:PyCT_algo}
\footnotesize
\KwIn{Classifier $C_M$, selected inputs $\Phi$, protected attributes $P\!A$, domain constraints $D$, resource limit}
\KwOut{Witness$(\varphi,\varphi')$, NoWitnessFound, or Timeout/Inconclusive}
\ForEach{seed input $\varphi$ in $\Phi$}{%
    Initialize $Q, T$ \tcp*{Both are empty at the beginning}
    $Q, T \gets$ \texttt{Exploration}($C_M, \varphi, P\!A, D, Q, T$) \tcp*{Explore paths induced by $\varphi$ (see~\cite{yu2024constraint})}
    \While{$Q$ is not empty}{%
        \If{the resource limit is reached}{%
            \Return Timeout/Inconclusive\;
        }
        $\phi \gets Q.\text{dequeue()}$ \tcp*{$\phi$ is the next path constraint}
        \If{$\phi \wedge D$ has a solution}{%
            $\varphi' \gets \text{result from SMT Solver}$ \tcp*{$\varphi'$ enables a new path}
            \If{$C_M(\varphi) \neq C_M(\varphi')$}{%
                \Return Witness$(\varphi, \varphi')$ \tcp*{An unfairness witness is found}
            }
            \Else{%
                $Q, T \gets$ \texttt{Exploration}($C_M, \varphi', P\!A, D, Q, T$) \tcp*{Explore paths induced by $\varphi'$}
            }
        }
    }
}
\Return NoWitnessFound \tcp*{All selected inputs have been processed}
\end{algorithm}

  In this section, we detail our methodology for evaluating individual fairness in DNNs through discriminatory instance checking.
  Our objective is to identify discriminatory instances for a pre-trained model, serving as evidence of the model's unfairness. We denote the attributes of the model input as $A = \{A_1, A_2, . . . , A_n\}$. Each attribute $A_i$ is associated with a domain $I_i$. The input domain is $D = I_1 \times I_2 \times \cdot \cdot \cdot \times I_n$, representing all possible combinations of the attribute values. We use $P\!A\subset A$ to represent the set of protected attributes, and use $N\!P\!A = A \setminus P\!A$ to denote the set of non-protected attributes. We distinguish the neural score function $M$ from the induced classifier $C_M$, where $C_M(x)$ denotes the predicted label obtained by applying the model's decision rule to $M(x)$.
    \begin{definition}[Protected-attribute counterfactual unfairness witness \citep{zhang2020white, li2023faire}] \label{def:def DI}
        Let $p$ denote the protected attributes and $n$ denote the non-protected attributes. A classifier $C_M$ has a \emph{protected-attribute counterfactual unfairness witness} with respect to $P\!A$ if there exist valid inputs $(p,n),(p',n) \in D$ such that $p \ne p'$ and $C_M(p,n) \ne C_M(p',n)$.
        For multiple protected attributes, $p \ne p'$ denotes that at least one protected coordinate differs while all non-protected coordinates are held fixed. We call the corresponding pair $((p,n),(p',n))$ an \emph{unfairness witness}; equivalently, $(p,n)$ is a discriminatory instance.
    \end{definition}

  A \emph{Deep Neural Network (DNN)} consists of an input layer, multiple hidden layers, and an output layer. Neurons in each layer connect to those in the adjacent layer through weighted connections, enabling information extraction and transformation. Frequently used activation functions include Rectified Linear Unit (ReLU), Sigmoid, Hyperbolic Tangent (Tanh), and Softmax. The increased depth and complexity of DNNs make them particularly effective for advanced tasks such as image processing, computer vision, and natural language processing.
  PyCT is engineered to parse and simulate the operations of network models embedded within Python programs, which includes implementing commonly used activation functions such as ReLU and Sigmoid. As a result, PyCT can analyze neural network models and Python programs in an integrated and coherent manner.

  % In practical applications, the ReLU activation function is widely adopted due to its superior convergence properties, efficient computation, and overall performance. Therefore, in this study, we employed fully connected networks based on ReLU activation, consistent with prior research efforts \citep{zhang2020white, zhang2021efficient, biswas2023fairify}. The mathematical expression of ReLU is given by Eq.(\ref{eq:relu}).
  %   \begin{equation} \label{eq:relu}
  %   \begin{aligned}[b]
  %       \text{ReLU}(x)=
  %       \begin{cases}
  %       0 &\text{if } x \leq 0 \text{ ; inactive neuron}\\
  %       x &\text{if } x > 0 \text{ ; active neuron}
  %       \end{cases}
  %   \end{aligned}
  %   \end{equation}

%\subsection{Example Execution of PyCT}
\begin{figure*}[t]
\centering
\includegraphics[width=\textwidth]{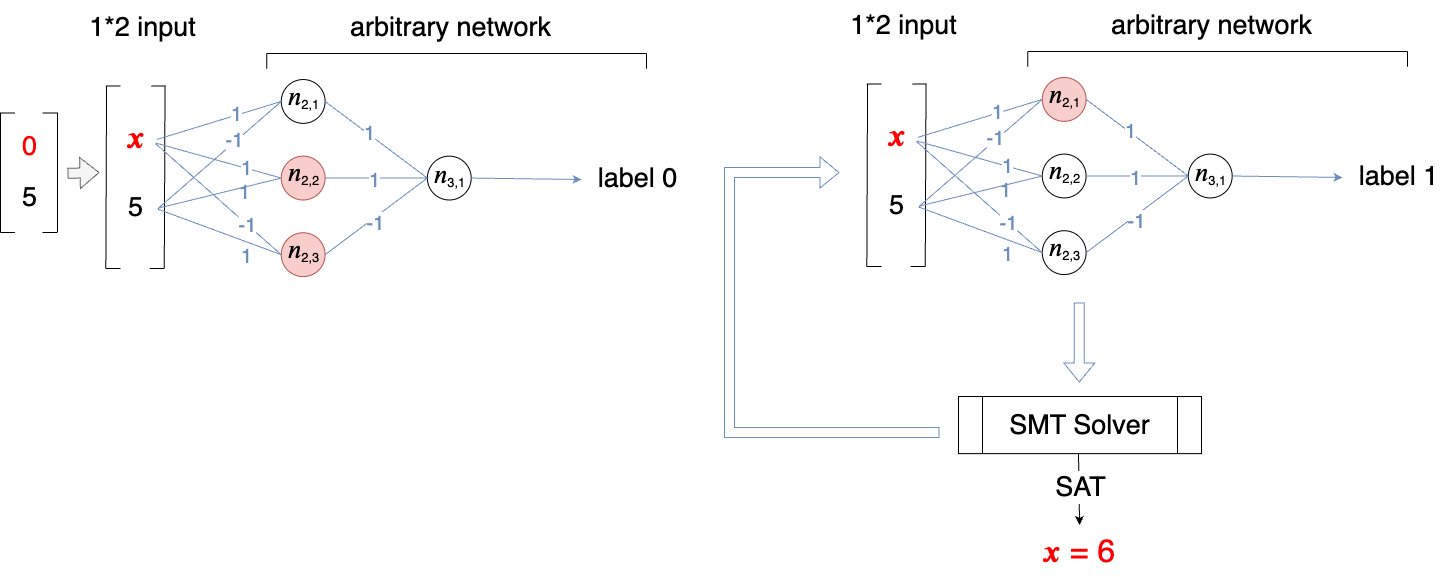}
\caption{A 3-layer DNN where the first attribute is protected (left). For this DNN and an input $\varphi$: [\textcolor{red}{0}, 5], PyCT identifies another input $\varphi'$: [\textcolor{red}{6}, 5] with a different model output (right), which indicates that $\varphi$ is a discriminatory instance for the DNN.
}
\label{fig:dis_running_ex}
\end{figure*}

  Figure~\ref{fig:dis_running_ex} illustrates how PyCT checks if a given input is discriminatory for a DNN. The example depicts a DNN with a 1$\times$2 input, undergoing a ReLU operation followed by a Sigmoid function for binary classification. A sample $[0,5]$ is given for discrimination evaluation, with the first attribute designated as protected (marked in red). %Additionally, the range of attributes in each instance is set $x$ from 0 to 10.
  We make the protected attribute value 0 a concolic variable $(0, x)$, which allows the perturbation of $x$ to identify attribute values that might alter the model's output. 
  The core algorithm for testing discriminatory instances is presented in Algorithm~\ref{alg:PyCT_algo}. Essentially,
  PyCT maintains a tree $T$ to track the path constraints associated with all explored network paths. It also manages a queue $Q$ containing formulas whose solutions correspond to input values that guarantee coverage of previously unexplored network paths.
  PyCT employs an SMT solver \citep{de2008z3,barbosa2022cvc5} to solve these formulas and find new test inputs that satisfy previously unexplored branch conditions. %The constraints provided to the solver are assertion statements, as illustrated in Listing~\ref{list:assert_1}.    
  In this example, the solver identifies a solution $x = 6$ for the perturbed input, leading to a new test case $[6,5]$. Feeding this case into the model changes its output from 0 to 1. Consequently, $\varphi$ is a discriminatory instance and we conclude that the DNN is unfair. 

\section{Witness Search and Conditional Verification with PyFair}
\label{c:model_fairness}
\begin{figure}
\centering
\includegraphics[width=\linewidth]{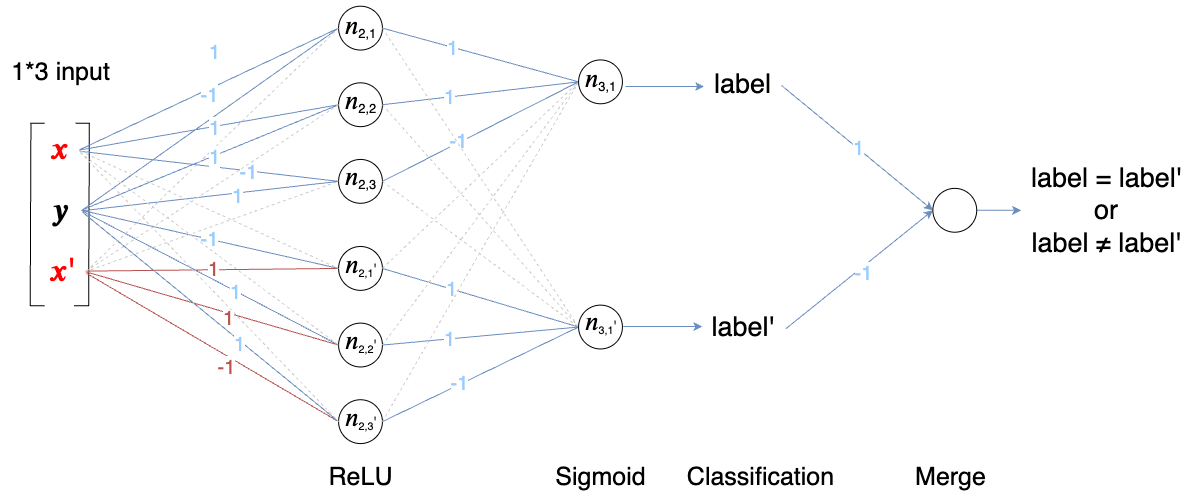}
\caption{\label{fig:DoubleDNN} The 2-DNN obtained from the DNN in Figure~\ref{fig:dis_running_ex}
}
\end{figure}

\begin{figure*}[t]
\centering
\includegraphics[width=\textwidth]{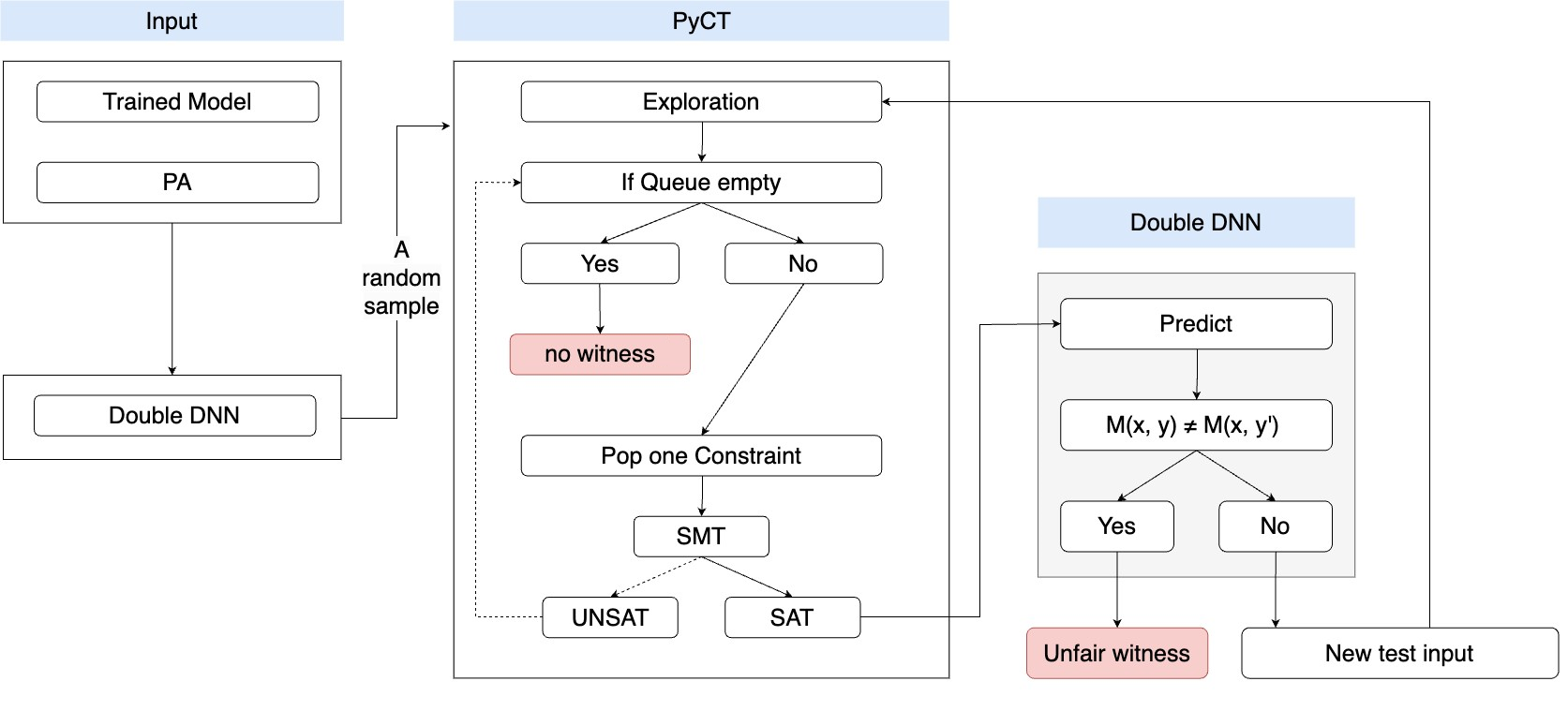}
\caption{An overview of the PyFair framework}
\label{fig:D-DNN_overview}
\end{figure*}

Model fairness checking searches the paired-input domain for instances $\varphi$ and $\varphi'$ that are identical in their NPAs but differ in their PAs and classifier labels.
As discussed in the previous section, PyCT can be directly applied to check model fairness. However, since PyCT can only check one instance at a time, it faces limitations when dealing with infinite input domains: even after testing numerous samples without detecting discrimination, PyCT often cannot conclusively establish model fairness.
To address this limitation, we propose a framework, named PyFair, to search for protected-attribute counterfactual unfairness witnesses and, under exact encoding and exhaustive exploration, verify their absence within the encoded domain. This framework essentially extends PyCT with an innovative data structure called the Dual DNN.
%PyFair extends the capabilities of PyCT, enabling a more thorough and systematic exploration of the model's behavior across its entire input domain. %thus offering a more robust approach to fairness verification.

\subsection{The Dual-DNN Architecture}

A \emph{Dual DNN (2-DNN)} is a DNN with three types of attributes \textsf{PA}, \textsf{NPA}, and \textsf{PA'}, where \textsf{PA'} duplicates \textsf{PA}. A 2-DNN $\widetilde{M}$ is constructed by creating two copies of a given DNN $M$, such that one takes an input on \textsf{PA} and \textsf{NPA}, and the other takes an input on \textsf{PA'} and \textsf{NPA}. The outputs of these two DNNs are combined by comparing the induced classifier labels.
As an illustration, consider the 2-DNN $\widetilde{M}$ in Figure~\ref{fig:DoubleDNN}, which is derived from the DNN $M$ in Figure~\ref{fig:dis_running_ex}. Given a (symbolic) input instance $(x,y)$ with $x\in$ \textsf{PA} and $y\in$ \textsf{NPA}, the 2-DNN transforms it to $(x, y, x')$, passes $x,y$ to the nodes $n_{2,1},n_{2,2},n_{2,3},n_{3,1}$, and passes $x',y$ to the nodes $n_{2,1}',n_{2,2}',n_{2,3}',n_{3,1}'$. The 2-DNN's output is determined by
$\widetilde{M}(x,y,x') = 1$ if $C_M(x,y) \neq C_M(x',y)$, and $\widetilde{M}(x,y,x') = 0$ otherwise.
Subsequently, we can employ PyCT to explore $\widetilde{M}$ and identify a solution for $\widetilde{M}(x,y,x')=1$ with $x \neq x'$. If the encoded path space is exhausted without such a solution, the original classifier has no unfairness witness under the encoded semantics and domain constraints. Thus, we can transform the fairness checking problem of a DNN into a witness-search problem for a 2-DNN.

\begin{theorem}[Dual-network equivalence]
Let $C_M$ be a deterministic classifier over a valid input domain $D$. Let $p$ denote the protected attributes and $n$ the non-protected attributes. Define the paired-input domain
\[
D_2 = \{(p,n,p') \mid (p,n) \in D,\ (p',n) \in D,\ p \ne p'\}.
\]
Construct a dual classifier $\widetilde{M}$ such that
\[
\widetilde{M}(p,n,p') = 1 \quad \text{iff} \quad C_M(p,n) \ne C_M(p',n).
\]
Then $C_M$ has a protected-attribute counterfactual unfairness witness if and only if there exists $(p,n,p') \in D_2$ such that $\widetilde{M}(p,n,p') = 1$.
\end{theorem}

\begin{proof}
For the forward direction, suppose $C_M$ has a witness $(p,n),(p',n) \in D$ with $p \ne p'$ and $C_M(p,n) \ne C_M(p',n)$. By construction, $(p,n,p') \in D_2$ and $\widetilde{M}(p,n,p')=1$. For the reverse direction, suppose there exists $(p,n,p') \in D_2$ with $\widetilde{M}(p,n,p')=1$. The definition of $D_2$ gives $(p,n),(p',n) \in D$ with $p \ne p'$, and the definition of $\widetilde{M}$ gives $C_M(p,n) \ne C_M(p',n)$. Thus the corresponding pair is an unfairness witness for $C_M$.
\end{proof}

\begin{corollary}[Conditional verification]
Assume that the classifier decision predicate, all activation constraints relevant to the decision, and the paired-input domain $D_2$ are encoded exactly in a sound and complete SMT theory. If PyFair exhaustively explores all feasible path regions of the dual network and every region is either inconsistent or satisfies $\widetilde{M}(p,n,p')=0$, then no protected-attribute counterfactual unfairness witness exists within the encoded domain.
\end{corollary}

Runs that terminate before exhaustive path exploration, for example because of a timeout, do not satisfy the premise of this corollary and are therefore reported as inconclusive rather than as verification results.

Algorithm~\ref{alg:Construct_double_dnn} implements the following block construction for the layer representation used by PyFair. Let the original network be written as a sequence of affine maps and activations,
\[
z^{k+1} = \sigma_k(W^k z^k + b^k).
\]
For layers after the input, the dual network uses block diagonal matrices:
\[
\widetilde{W}^k =
\begin{bmatrix}
W^k & 0 \\
0 & W^k
\end{bmatrix},
\qquad
\widetilde{b}^k =
\begin{bmatrix}
b^k \\
b^k
\end{bmatrix}.
\]
For the first affine layer, partition the original first-layer matrix as $W^0 = [W^0_{P}\; W^0_{N}]$, where the columns correspond to protected and non-protected attributes. With paired inputs ordered as $(p,n,p')$, the first dual affine map is
\[
\widetilde{W}^0 =
\begin{bmatrix}
W^0_{P} & W^0_{N} & 0 \\
0 & W^0_{N} & W^0_{P}
\end{bmatrix},
\qquad
\widetilde{b}^0 =
\begin{bmatrix}
b^0 \\
b^0
\end{bmatrix}.
\]
The final comparison node is $\widetilde{M}(p,n,p') = \mathbf{1}[C_M(p,n) \ne C_M(p',n)]$. For binary classifiers, $C_M$ is obtained by the classifier's decision threshold; for multiclass classifiers, $C_M$ is obtained by the classifier's label-selection rule such as argmax when that is the encoded decision rule.

\begin{algorithm}[t]
\caption{Dual-DNN block construction}
\label{alg:Construct_double_dnn}
\footnotesize
\KwIn{Classifier network $M$ with affine layers $(W^k,b^k)$ and protected-attribute columns $P\!A$}
\KwOut{Dual classifier $\widetilde{M}$ over paired input $(p,n,p')$}
Partition the original input as $(p,n)$ and $W^0$ as $[W^0_P\; W^0_N]$\;
Set $\widetilde{W}^0 \gets \begin{bmatrix} W^0_P & W^0_N & 0 \\ 0 & W^0_N & W^0_P \end{bmatrix}$ and $\widetilde{b}^0 \gets \begin{bmatrix} b^0 \\ b^0 \end{bmatrix}$\;
\ForEach{hidden affine layer $k \geq 1$}{%
    Set $\widetilde{W}^k \gets \begin{bmatrix} W^k & 0 \\ 0 & W^k \end{bmatrix}$ and $\widetilde{b}^k \gets \begin{bmatrix} b^k \\ b^k \end{bmatrix}$\;
    Apply the same activation $\sigma_k$ independently to both copies\;
}
Define the final comparison node as $\widetilde{M}(p,n,p') \gets \mathbf{1}[C_M(p,n) \ne C_M(p',n)]$\;
\Return{$\widetilde{M}$}\;
\end{algorithm}

\subsection{Decision semantics and SMT encoding}

PyFair's solver results apply to the mathematical classifier encoded into SMT. For networks whose relevant decision predicates can be represented using linear arithmetic, such as ReLU networks followed by linear threshold or argmax decisions over logits, the path constraints are encoded in linear real arithmetic. If the implemented model contains numerical operations such as Sigmoid or Softmax, PyFair encodes the induced decision predicate used for classification rather than relying on an exact encoding of transcendental functions, unless such an exact encoding is explicitly provided. Consequently, a verification result is a statement about the encoded decision semantics and domain constraints, not automatically about every floating-point implementation detail.

\subsection{Input-domain constraints}

The paired-input domain $D_2$ must include the validity constraints of the original feature space. Continuous attributes are constrained by their admissible lower and upper bounds. Integer and categorical attributes require finite-domain constraints, and one-hot encoded categories require exactly-one constraints. When features are normalized before being supplied to the network, these constraints are applied consistently in the normalized representation or are transformed from the original feature space into the normalized coordinates. For multiple protected attributes, the constraint $p \ne p'$ is encoded as a disjunction stating that at least one protected coordinate differs, while all non-protected coordinates are constrained to be equal across the two copies. The validity of solver-generated witnesses depends on including these constraints; otherwise, a solver may return feature combinations that are outside the intended data domain.

\subsection{The PyFair Framework}

We outline the framework of PyFair in Figure~\ref{fig:D-DNN_overview}. PyFair takes a DNN model and a set of PAs as input. It first constructs a 2-DNN based on the model and the PAs. Starting from a random concolic input, PyFair either finds an unfairness witness, reports no witness found for the explored search, or returns an inconclusive result when resource limits stop exploration. A no-witness result is a verification result only under the exact-encoding and exhaustive-exploration conditions stated in the conditional verification corollary.

\paragraph{Finding an unfairness witness with PyFair.}
\label{unfair_running_example}

\begin{figure*}[t]
\centering
\includegraphics[width=\textwidth]{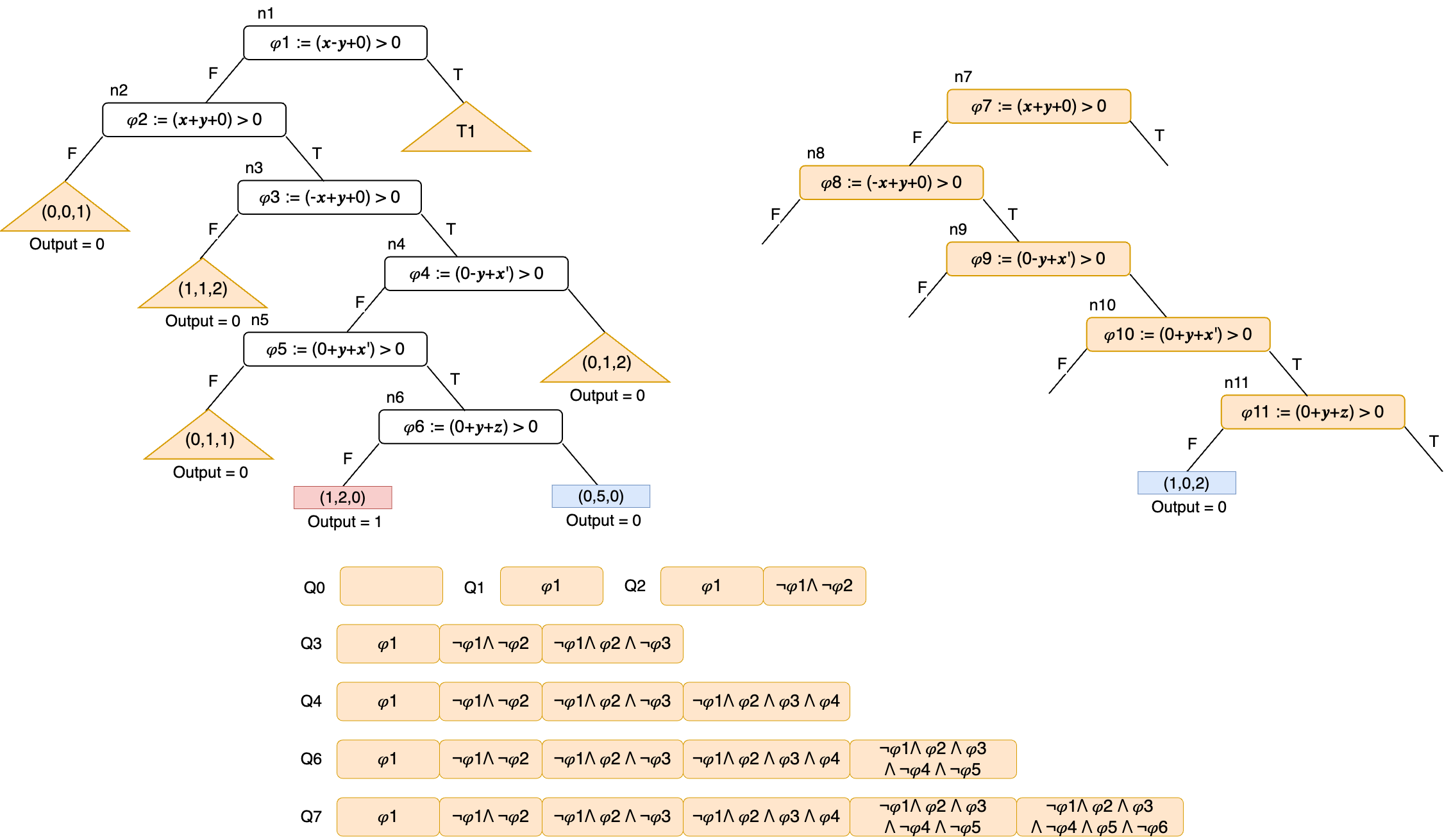}
\caption{Tree $T$ (top left), Tree $T'$ (top right), and Queue $Q$ (below)}
\label{fig:unfair_example}
\end{figure*}

  As a detailed example, we describe the execution of PyFair on the 2-DNN in Figure~\ref{fig:DoubleDNN} and the input $[0, 5, 0]$. As mentioned earlier, the concolic tester maintains a tree $T$ and a queue $Q$ to explore previously unexamined network branches.
  %The concolic tester systematically evaluates each constraint in $Q$ by generating test inputs. Adding assertions to define the range limits for these test inputs is important. In this example, the value range for $(x, y, x')$ is set from 0 to 10. 
  Figure~\ref{fig:unfair_example} depicts the states of $T$ and $Q$ during discriminatory instance checking.

  \emph{The $1^{st}$ iteration:} We set three input as concolic variable: $(0,x)$, $(5,y)$, and $(0,x')$. After computing the weighted sum, the first hidden node $n_{2,1}$ remains a concolic variable $(-5, x-y)$, computed as $0*1+5*(-1)+0*0=-5$ for the concrete value and $x*1+y*(-1)+x'*0=x-y$ for the symbolic expression. A node $n_{2,1}$ with label $\varphi_1=(x-y)>0$ is inserted into $T$ as a root. The remaining hidden nodes $n_{2,1},n_{2,2},n_{2,3},n_{3,1},n_{2,1}',n_{2,2}',n_{2,3}',n_{3,1}'$ are computed in similar manner. The output vector is computed using the absolute difference between the variables $label$ and $label'$ to determine if the two labels are the same. Since our input is $(0,5,0)$, we collect the input along with its output as $x=0$, $y=5$, $x'=0$, and $label=label'$.
    % \begin{python}[
    %     float,floatplacement=ht,
    %     caption={Constraint for $\varphi_1$.},
    %     label=list:assert_2,
    %     escapechar=|]
    % (set-logic ALL)
    % (declare-const x Int)
    % (declare-const y Int)
    % (declare-const x' Int)
    % (assert (not (<= (+ 0.0 (+ (* y (- 1.0)) (* x 1.0))) 0)))
    % (assert (and (<= x 10) (>= x 0)))
    % (assert (and (<= y 10) (>= y 0)))
    % (assert (and (<= x' 10) (>= x' 0)))
    % (assert (distinct x x')
    % (check-sat)
    % (get-value (x))
    % (get-value (y))
    % (get-value (x'))
    % \end{python}

  \emph{The $2^{nd}$ iteration:} We give the constraints dequeued from $Q$ to the SMT solver, which
  %In this case, line 5 addresses the unexplored path constraint, while lines 6 to 8 define the range of the symbolic variable (assumed to be 0-10). Line 9 includes an additional assertion to define our fairness testing: the $x$ and $x'$ values in the unfairness witness should not be identical.
  provides a solution $x=1, y=0, x'=2$ as a new test input. We repeat this procedure to gather more branches, as illustrated by Tree $T'$ in Figure~\ref{fig:unfair_example}. %This figure shows how this new test input collects branch conditions following the previously discussed method.
  Finally, the Sigmoid function produces the same prediction as the original label. Thus, we continue dequeuing constraints from $Q$.
  %
  %This time, the solver gives a solution $x=1, y=2, x'=0$. We repeat the procedure as we did before to collect more branches (illustrated in the right of Figure~\ref{fig:unfair_ex_tree}). Finally, the Sigmoid function gives the prediction class 0, which is the same as the origin label, so we continuously dequeue constraints from $Q$.

  \emph{The $3^{rd}\sim 6^{th}$ iterations:} Constraints $\neg\varphi_1 \wedge \neg\varphi_2$, $\neg\varphi_1 \wedge \varphi_2 \wedge \neg\varphi_3$, etc., are dequeued from $Q$ in order by exploring new test inputs. This process terminates when the queue $Q$ is empty, or the output label is 1 (indicating the discovery of an unfairness witness). Since the outputs remain unchanged for these inputs, the process continues.

  \emph{The $7^{th}$ iteration:} In this iteration, the constraint $\neg\varphi_1 \wedge \varphi_2 \wedge \varphi_3 \wedge \neg\varphi_4 \wedge \varphi_5 \wedge \neg\varphi_6$ is dequeued from $Q$. This time, the test input $x=1, y=2, x'=0$ yields a different model prediction. We add a node with this input and the prediction label $Output = 1$ as the left child of \textsf{n6}.
  The process terminates here and reports unfairness for the model. The witness of unfairness is $((1,2), (0,2))$.

\paragraph{Conditional verification with PyFair.}
  We proceed to demonstrate how PyFair can verify the absence of witnesses under the encoding for an artificially fair model. To obtain this model, we modify the DNN in Figure~\ref{fig:dis_running_ex} by setting the outgoing weights of the PA to zero. With this setup, the PA value does not influence the output value. %We conduct model fairness checking on this model with an initial test input $[0,5,0]$.

% \begin{python}[
%         float,floatplacement=ht,
%         caption={Constraint for $\varphi_1$.},
%         label=list:assert_3,
%         escapechar=|]
%     (assert (not (<= (+ 0.0 (+ (* y (-1.0)) 0)) 0)))
%     (assert (and (<= x 5) (>= x 0)))
%     (assert (and (<= y 5) (>= y 0)))
%     (assert (and (<= x' 5) (>= x' 0)))
%     (assert (distinct x x')
% \end{python}
    % \begin{python}[
    %     float,floatplacement=ht,
    %     caption={Constraint for $\neg\varphi_1 \wedge \neg\varphi_2$.},
    %     label=list:assert_4,
    %     escapechar=|]
    % (set-logic ALL)
    % (declare-const x Int)
    % (declare-const y Int)
    % (declare-const x' Int)
    % (assert (<= (+ 0.0 (+ (* y (-1.0)) 0)) 0))
    % (assert (<= (+ 0.0 (+ (* y 1.0) 0)) 0))
    % (assert (and (<= x 5) (>= x 0)))
    % (assert (and (<= y 5) (>= y 0)))
    % (assert (and (<= x' 5) (>= x' 0)))
    % (assert (distinct x x')
    % (check-sat)
    % (get-value (x))
    % (get-value (y))
    % (get-value (x'))
    % \end{python}
    % \begin{python}[
    %     float,floatplacement=ht,
    %     caption={Constraint for $\neg\varphi_1 \wedge \varphi_2 \wedge \neg\varphi_3$.},
    %     label=list:assert_5,
    %     escapechar=|]
    % (set-logic ALL)
    % (declare-const x Int)
    % (declare-const y Int)
    % (declare-const x' Int)
    % (assert (<= (+ 0.0 (+ (* y (-1.0)) 0)) 0))
    % (assert (not (<= (+ 0.0 (+ (* y 1.0) 0)) 0)))
    % (assert (<= (+ 0.0 (+ (* y 1.0) 0)) 0))
    % (assert (and (<= x 5) (>= x 0)))
    % (assert (and (<= y 5) (>= y 0)))
    % (assert (and (<= x' 5) (>= x' 0)))
    % (assert (distinct x x')
    % (check-sat)
    % (get-value (x))
    % (get-value (y))
    % (get-value (x'))
    % \end{python}

\begin{figure*}[t]
\centering
\includegraphics[width=\textwidth]{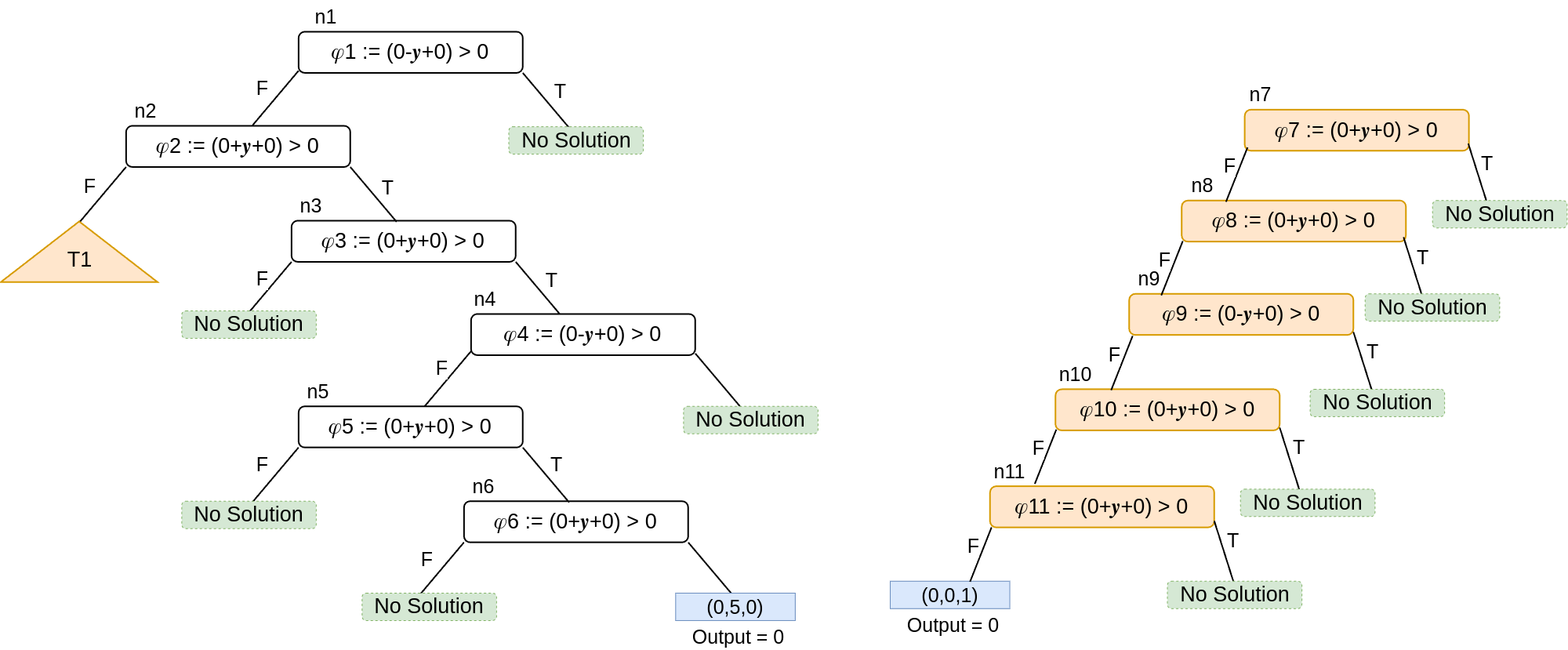}
\caption{Tree $T$ (left) and Tree $T'$ (right)}
\label{fig:fair_example}
\end{figure*}

  \emph{The $1^{st}$ iteration:} We use $[0,5,0]$ as the initial input to explore the branch conditions. The $T$ and $Q$ collected by PyFair are displayed in Figure~\ref{fig:fair_example}.

  \emph{The $2^{nd}$ iteration:} In this iteration, we proceed to examine the branches starting from the front of $Q$ (Q7 in Figure~\ref{fig:fair_example}). 
  %The constraints ($\varphi_1$) appear as shown in List~\ref{list:assert_3}, where line 5 specifies the constraints for exploring the desired branch, and lines 6-9 handle input range limitations. 
  The SMT solver cannot find a solution, so we label the corresponding path as ``no solution'' (i.e., at the left branch of \textsf{n6} in Tree $T$) in Figure~\ref{fig:fair_example}.

  \emph{The $3^{rd}$ iteration:} In this step, we address the constraint $\neg\varphi_1 \wedge \neg\varphi_2$. The SMT solver generates a new test input $[0,0,1]$ and repeats the exploration process. The path condition explored by this third test input is depicted by Tree $T'$ in Figure~\ref{fig:fair_example}. The prediction label for $[0,0,1]$ remains unchanged from the original. Consequently, we insert the values $x=0, y=0, x'=1, Output=0$ into the left branch of \textsf{n11}.

  \emph{The $4^{th}$ iteration:} The process continues by dequeuing from $Q$. In this iteration, $\neg\varphi_1 \wedge \varphi_2 \wedge \neg\varphi_3$ has no solution, so we label ``no solution'' on the corresponding branch in the tree, similar to what was done in iteration 2.
  
  \emph{The $5^{th}\sim 12^{th}$ iterations:} Similarly, for the 5th to 12th iterations, PyFair continues by dequeuing $Q$ to resolve constraints. Since the SMT solver returns UNSAT, indicating no solutions for each branch, we label ``no solution'' on the branches in Figure~\ref{fig:fair_example}.
  
  Once all path constraints are resolved (i.e., $Q$ becomes empty), PyFair reports that no unfairness witness can be identified for the network. Since the network and paired-input domain are exactly encoded in SMT in this example, this outcome verifies fairness under the encoding with respect to the given PAs.

\section{Evaluation}
\label{c:evaluation}
    In our evaluation, we conduct protected-attribute counterfactual witness search and conditional verification experiments for 25 models. We adopt a methodology similar to Fairify~\citep{biswas2023fairify} and employ benchmark models from Fairify and previous studies~\citep{zhang2021efficient}. These neural networks are fully connected and use ReLU and Sigmoid operations in the model implementation. RQ1 evaluates the underlying PyCT-based single-instance witness search on the original benchmark models. RQ2 and RQ4 evaluate the dual-network PyFair construction in multi-protected-attribute and verification settings. We also test models generated by existing bias mitigation techniques like ADF and EIDIG~\citep{zhang2020white, zhang2021efficient}, evaluating how effectively PyFair can identify discriminatory instances in retrained models (RQ3).
    Finally, we investigate PyFair's ability to verify the absence of witnesses under the encoded semantics for artificially fair models (RQ4). % We manually modified the benchmarks by setting the outgoing weights from PAs to zero, ensuring that attribute values do not influence the output. 
  % Finally, in addition to setting the PA outgoing weight to zero, we modified the output layer activation function to use a weighted sum and employed zero as the threshold for classification. This adjustment was made because PyCT can only approximate values when dealing with Sigmoid calculations and cannot guarantee fairness. By implementing these changes, we aim to evaluate how effectively model fairness checking performs in verifying models that are designed to be entirely fair (RQ4).
  
% \section{Research Questions}
%   In this study, we answered four research questions for evaluation:
  
%   \textbf{\emph{RQ1:}} What is the performance of discriminatory instance checking using PyCT on PA?

%   \textbf{\emph{RQ2:}} What is the performance of PyCT to conduct model fairness checking on multiple PA?

%   \textbf{\emph{RQ3:}} How effective is model fairness checking in the fairness improvement model by ADF and EIDIG?
  
%   \textbf{\emph{RQ4:}} How does model fairness checking perform in demonstrating the model's fairness?
        We will use the following indicators in the tables:
        \textbf{UW} stands for whether an unfairness witness was found, with Y indicating that a discriminatory instance or paired-input witness is found, N indicating that no witness was found during the run, and Unk meaning no conclusion within the time limit. A value of UW=N is not automatically a fairness proof.
        \textbf{FQ} denotes the number of frontier/path obligations generated after the initial symbolic pass. It is an operational measure of the explored search frontier and may correlate with model branching complexity, but it is not itself a general model-size metric.
        \textbf{\#test} is the total number of seed or generated inputs tested within the reported run.
        \textbf{\#sat} and \textbf{\#unsat} count solver outcomes during the run. These counts indicate the amount of solver exploration performed, but they imply exhaustive coverage only when the run is explicitly marked as exhausted.
        \textbf{Time (s)} is the execution time in seconds as recorded by the experimental scripts. Timeout entries indicate runs terminated by the configured resource limit. The Fairify and RQ4 timeout entries use an 1800-second limit; RQ1 PyCT times are reported for the PyCT runs over checked seeds and subprocesses and therefore are not identical to Fairify's timeout metric.
        \textbf{Bias(\%)} is employed to assess model fairness by determining the proportion of individual discriminatory instances within the dataset, and a lower percentage indicates greater fairness of the model.\footnote{More precisely, we repeatedly sampled 100 random inputs from the dataset and altered the PA value at random. If changing PA results in a classification shift, it is considered a discriminatory instance. We then compute the ratio of discriminatory inputs over 100 rounds. This calculation method is adapted from AEQUITAS~\citep{udeshi2018automated}.} %According to statistical theory, the average ratio will converge to the expected value.}
        \textbf{Fair} is used only for the artificially fair models in RQ4: Y means verified fair under the encoding, N means a witness is found, and Unk means timeout or otherwise inconclusive.

\paragraph{RQ1: How does PyCT single-instance witness search perform on a single PA?}
  In this experiment, we use 1500 benchmark instances as seed inputs for PyCT. We evenly distribute these samples among 30 subprocesses for parallel processing. Each subprocess independently performs discriminatory instance checking on its assigned 50 samples.
  %If any subprocess finds a discriminatory instance, the checking terminates and the time is recorded. 
  We present the results in Table~\ref{tab:rq1}, where column ``UW'' indicates whether PyCT or Fairify found an unfairness witness in the reported run.
  %If PyCT detects discrimination in the random 1500 samples, we mark Y in the column ``UW'' and mark N otherwise.
  %PyCT sets the PA as a concolic variable and individually performs discriminatory instance checking for each sample. 

\begin{table}[ht]
    \caption{RQ1 witness-finding outcomes for PyCT single-instance search and Fairify. In the UW columns, Y means an unfairness witness was found, while N means no witness was found during the run.}
    \label{tab:rq1}
    \centering
    \small
    \setlength{\tabcolsep}{5pt}
    \textbf{Protected attribute: Race}\\[0.35em]
    \begin{tabular}{@{}l r c c r r r r r@{}}
    \toprule
    & & Fairify & \multicolumn{6}{c}{PyCT}\\
    \cmidrule(lr){3-3}\cmidrule(l){4-9}
    Model & Bias(\%) & UW & UW & FQ & \#test & \#sat & \#unsat & Time (s) \\
    \midrule
    AC1 &2.11 &Y &Y &24 &10 &3 &256 &33.46\\
    AC2 &0.8 &Y &Y &100 &2 &1 &117 &33.55\\
    AC3 &1.84 &Y &Y &50 &5 &3 &244 &33.56\\
    AC4 &1.06 &N &Y &200 &3 &1 &539 &408.45\\
    AC5 &2.48 &Y &Y &128 &1 &1 &79 &106.82\\
    AC6 &0.86 &Y &Y &24 &11 &6 &321 &467.13\\
    AC7 &0.78 &N &Y &124 &71 &60 &11082 &2697.17\\
    AC8 &1.89 &Y &Y &10 &95 &25 &950 &175.38\\
    AC9 &2.01 &Y &Y &12 &30 &9 &348 &110.51\\
    AC10 &1.54 &Y &Y &20 &11 &1 &213 &33.61\\
    AC11 &0.91 &N &Y &40 &7 &1 &250 &1684.83\\
    AC12 &1.54 &N &Y &45 &5 &1 &179 &469.69\\
    \bottomrule
    \end{tabular}

    \vspace{0.9em}
    \textbf{Protected attribute: Age}\\[0.35em]
    \begin{tabular}{@{}l r c c r r r r r@{}}
    \toprule
    & & Fairify & \multicolumn{6}{c}{PyCT}\\
    \cmidrule(lr){3-3}\cmidrule(l){4-9}
    Model & Bias(\%) & UW & UW & FQ & \#test & \#sat & \#unsat & Time (s) \\
    \midrule
    BM1 &0 &Y &Y &80 &138 &5 &11036 &1053.78\\
    BM2 &0 &Y &Y &48 &13 &2 &589 &33.46\\
    BM3 &0 &Y &Y &100 &58 &1 &5705 &622.20\\
    BM4 &0 &Y &Y &300 &590 &20 &147165 &36028.09\\
    BM5 &0.45 &Y &Y &32 &209 &6 &6512 &388.63\\
    BM6 &0.88 &Y &Y &18 &220 &13 &3822 &332.68\\
    BM7 &1.42 &Y &Y &128 &68 &3 &8366 &1186.26\\
    BM8 &0.52 &N &Y &124 &71 &15 &8976 &2400.23\\
    \bottomrule
    \end{tabular}

    \vspace{0.9em}
    \textbf{Protected attribute: Sex}\\[0.35em]
    \begin{tabular}{@{}l r c c r r r r r@{}}
    \toprule
    & & Fairify & \multicolumn{6}{c}{PyCT}\\
    \cmidrule(lr){3-3}\cmidrule(l){4-9}
    Model & Bias(\%) & UW & UW & FQ & \#test & \#sat & \#unsat & Time (s) \\
    \midrule
    GC1 &0.98 &Y &Y &50 &7 &2 &347 &33.96\\
    GC2 &2.07 &Y &Y &100 &2 &1 &145 &68.03\\
    GC3 &1.9 &Y &Y &9 &17 &1 &147 &42.73\\
    GC4 &0 &Y &N &10 &1500 &0 &15000 &52217.74\\
    GC5 &0 &N &N &124 &1443 &6 &169103 &60809.37\\
    \bottomrule
    \end{tabular}
\end{table}

\begin{table}[t]
\centering
\caption{Summary of witness-finding outcomes in RQ1, derived from Table~\ref{tab:rq1}.}
\label{tab:rq1-summary}
\begin{tabular}{lr}
\toprule
Outcome & Count \\
\midrule
Models evaluated & 25 \\
Witness found by PyCT & 23 \\
Witness found by Fairify & 19 \\
Witness found by PyCT but not Fairify & 5 \\
Witness found by Fairify but not PyCT & 1 \\
\bottomrule
\end{tabular}
\end{table}
  
  We compare our tool with Fairify~\citep{biswas2023fairify}, a state-of-the-art constraint-based fairness verifier, over the same PAs on the same datasets.
  %This alignment ensures that our comparisons are consistent and meaningful.
  %, allowing us to accurately evaluate the relative effectiveness of PyCT in identifying discriminatory instances and verifying model fairness.
  Fairify identifies discriminatory instances for 19 models with a timeout of 1800 seconds, as shown in column ``Fairify UW''. Detailed information from PyCT is also provided: ``\#test'' indicates the total number of instances tested on the model, and ``\#sat'' and ``\#unsat'' count feasible and infeasible solver queries during the run.
  %PyCT can provide more detailed information compared to other tools in identifying discriminatory instances.
  For instance, in the case of AC5, PyCT identifies its discriminatory instance right after the first test sample. It locates this instance after making 80 branch exploration attempts (the sum of \#sat and \#unsat). 
  The SAT outcomes indicate the number of successful explorations, while the UNSAT ones indicate the visited hidden nodes that do not impact the output.

  Compared with Fairify on the same protected attributes, PyCT finds witnesses in more of the listed benchmark models. The comparison is not uniformly favorable, since Fairify reports a witness for GC4 where PyCT does not. We therefore interpret the result as evidence of complementary witness-finding ability rather than as a uniform dominance claim. In some cases, PyCT takes a long time to test many samples, e.g., for GC4 and GC5, possibly due to insufficient seed diversity. In the subsequent experiment, we attempt to address this limitation using 2-DNN.

\paragraph{RQ2: How does PyFair perform in witness search with multiple PAs?}
  In contrast to vanilla PyCT, PyFair is capable of checking discriminatory instances with multiple PAs thanks to the use of 2-DNN. %of automatically generating unfair witness where a pair of instances with only the selected PAs differ while keeping the NPAs the same and the classification of such pairs is different.
  For dual\_ACs (i.e., the 2-DNN counterpart of the AC models), the runtime of checking multiple PAs is faster than checking a single PA in many tests. The ``FQ'' values of these tests describe the initial search frontier generated by the dual network. %This demonstrates that PyFair exhibits good scalability and adaptability in addressing complex fairness issues. 
  Moreover, the values of \#sat and \#unsat are generally lower than the corresponding PyCT runs in RQ1, indicating that PyFair can identify discriminatory instances with fewer solver outcomes in these cases. % exhibits a similar performance pattern on other models. %(in Figure~\ref{fig:RQ2_pyct_fc}).
  %\input{figures/RQ2_pyct_fc}

  % Furthermore, we compared the ability of PyCT and PyFair to find new test inputs. Figure~\ref{fig:RQ2_combine}~(1) shows the ratio of SAT and UNSAT solved by the SMT solver for \#test during the witness of unfairness in each AC model using PyCT. Figure~\ref{fig:RQ2_combine}~(2) presents the \#sat and \#unsat ratios during the witness of unfairness with PyFair. PyFair is more effective in identifying valid test inputs for testing the AC model when considering multiple PAs. This is demonstrated by the significantly higher number of SAT (shown in blue in Figure~\ref{fig:RQ2_combine}) compared to the UNSAT ones (orange in Figure~\ref{fig:RQ2_combine}).
  % %\input{figures/RQ2_combine}

\begin{table}[ht]
\centering
\caption{RQ2 witness-finding outcomes for PyFair on dual networks with multiple protected attributes. UW=Y means an unfairness witness was found; UW=N means no witness was found during the run and is not by itself a fairness proof.}
\label{tab:rq2}
\small
\setlength{\tabcolsep}{7pt}
\textbf{Protected attributes: Race, Age, Sex}\\[0.35em]
\begin{tabular}{@{}l c r r r r@{}}
\toprule
\multicolumn{6}{c}{PyFair}\\
\cmidrule(lr){1-6}
Model & UW & FQ & \#sat & \#unsat & Time (s)\\
\midrule
dual\_AC1 &Y &48 &36 &19 &143.06\\
dual\_AC2 &Y &200 &1 &0 &47.8\\
dual\_AC3 &Y &100 &4 &0 &148.78\\
dual\_AC4 &Y &400 &20 &1 &333.41\\
dual\_AC5 &Y &256 &4 &0 &146.78\\
dual\_AC6 &Y &48 &27 &25 &1380.37\\
dual\_AC7 &Y &248 &13 &3 &3078.09\\
dual\_AC8 &Y &20 &18 &1 &16.36\\
dual\_AC9 &Y &24 &9 &6 &11.94\\
dual\_AC10 &Y &40 &5 &0 &10.19\\
dual\_AC11 &Y &80 &27 &4 &1413.52\\
dual\_AC12 &Y &90 &6 &1 &1653.62\\
\bottomrule
\end{tabular}

\vspace{0.9em}
\textbf{Protected attributes: Age, Sex}\\[0.35em]
\begin{tabular}{@{}l c r r r r@{}}
\toprule
\multicolumn{6}{c}{PyFair}\\
\cmidrule(lr){1-6}
Model & UW & FQ & \#sat & \#unsat & Time (s)\\
\midrule
dual\_GC1 &Y &100 &166 &134 &41043.57\\
dual\_GC2 &Y &200 &6 &2 &171.08\\
dual\_GC3 &Y &18 &5 &1 &11.18\\
dual\_GC4 &N &20 &258 &1833 &1841.44\\
dual\_GC5 &N &248 &65 &46 &Timeout\\
\bottomrule
\end{tabular}
\end{table}

% \begin{table}[ht]
% \centering
% \caption{Checking Fairness of Biased Models}
% \label{Tab:pyct_double_dnn_result}
% \resizebox{0.55\columnwidth}{!}{
% \begin{tabular}[t]{ll|crrrr}
% \toprule
% & & \multicolumn{5}{c}{PyFair}\\
% PA & Model &UW &FQ &\#sat &\#unsat &Time\\
% \midrule
% % \multirow{12}{3.5em}{Race} &dual\_AC1 &Y &48 &154 &188 &1088.16\\
% % &dual\_AC2 &Y  &200 &35 &7 &709.74\\
% % &dual\_AC3 &Y &100 &333 &512 &9369.54\\
% % &dual\_AC4 &Y &400 &94 &322 &21202.29\\
% % &dual\_AC5 &Y &256 &279 &1317 &40139.54\\
% % &dual\_AC6 &Y &48 &105 &119 &929.05\\
% % &dual\_AC7 &Y &248 &44 &55 &5584.06\\
% % &dual\_AC8 &N &20 &288 &327 &Timeout\\
% % &dual\_AC9 &Y &24 &11 &14 &11.40\\
% % &dual\_AC10 &N &40 &1185 &4276 &Timeout\\
% % &dual\_AC11 &Y &80 &362 &422 &46613.67\\
% % &dual\_AC12 &Y &90 &13 &3 &6401.41 \\
% % \midrule
% \multirow{12}{3.5em}{Race, Age, Sex} &dual\_AC1 &Y &48 &36 &19 &143.06\\
% &dual\_AC2 &Y &200 &1 &0 &47.8\\
% &dual\_AC3 &Y &100 &4 &0 &148.78\\
% &dual\_AC4 &Y &400 &20 &1 &333.41\\
% &dual\_AC5 &Y &256 &4 &0 &146.78\\
% &dual\_AC6 &Y &48 &27 &25 &1380.37\\
% &dual\_AC7 &Y &248 &13 &3 &3078.09\\
% &dual\_AC8 &Y &20 &18 &1 &16.36\\
% &dual\_AC9 &Y &24 &9 &6 & 11.94\\
% &dual\_AC10 &Y &40 &5 &0 &10.19\\
% &dual\_AC11 &Y &80 &27 &4 &1413.52\\
% &dual\_AC12 &Y &90 &6 &1 &1653.62 \\
% \midrule
% \multirow{5}{3.5em}{Age, Sex} &dual\_GC1 &Y &100 &166 &134 &41043.57\\
% &dual\_GC2 &Y &200 &6 &2 &171.08\\
% &dual\_GC3 &Y &18 &5 &1 &11.18\\
% &dual\_GC4 &N &20 &258 &1833 &1841.44\\
% &dual\_GC5 &N &248 &65 &46 &Timeout\\
% \bottomrule
% \end{tabular}
% }
% \end{table}%

\begin{table}[t]
\centering
\caption{Summary of witness-finding outcomes in RQ2, derived from Table~\ref{tab:rq2}.}
\label{tab:rq2-summary}
\renewcommand{\arraystretch}{1.15}
\begin{tabular*}{0.78\textwidth}{@{\extracolsep{\fill}}lr@{}}
\toprule
Outcome & Count \\
\midrule
Dual models evaluated & 17 \\
Witness found by PyFair & 15 \\
No witness found without a timeout marker & 1 \\
Timeout/inconclusive & 1 \\
\bottomrule
\end{tabular*}
\end{table}

  Compared with vanilla PyCT, PyFair starts from one concolic seed and can generate paired inputs by solving path constraints over the dual network. For example, for dual\_GC4, PyFair records 258 SAT and 1833 UNSAT solver outcomes without finding a witness, while PyCT tests the GC4 model over 1500 seed inputs without identifying a new test case (see Table~\ref{tab:rq1}). 
  Although both approaches are inconclusive for this case, the dual-network run explores a larger constraint-guided search frontier. These results show PyFair's ability to handle multiple PAs simultaneously. On the other hand, PyFair still struggles with complex models like GC5, which indicates the challenges in thoroughly evaluating DNN fairness and the need for further optimization.

\paragraph{RQ3: How effective is PyFair in testing models retrained by ADF and EIDIG?}
  ADF and EIDIG~\citep{zhang2020white, zhang2021efficient} exploit discriminatory instances to augment the data and retrain the model. Both retraining methods involve randomly sampling 5\% of the generated discriminatory instances, relabeling them using majority voting~\citep{lam1997application}, and incorporating these instances into the original training set before retraining the model on the augmented dataset. The effectiveness of these methods is shown in the retrained-model block of Table~\ref{Tab:pyct_EIDIG}. As observed, models AC14, AC15, and AC16 exhibit lower Bias(\%) compared to AC13. The difference between AC15 and AC16 lies in the frequency of recalculating gradients and attribute contributions. In AC15, these calculations are updated every five iterations (EIDIG-5), whereas in AC16 they are not updated (EIDIG-$\infty$). 

\begin{table}[t]
\caption{PyFair outcomes on retrained biased models and artificially fair models. UW=Y means an unfairness witness was found. Fair=Y means verified fair under the encoded semantics, and Unk means timeout or inconclusive.}
\label{Tab:pyct_EIDIG}
\centering
\small
\setlength{\tabcolsep}{6pt}
\textbf{Retrained biased models; protected attributes: Race, Age, Sex}\\[0.35em]
\begin{tabular}{@{}l r c r r r r@{}}
\toprule
\multicolumn{7}{c}{PyFair}\\
\cmidrule(lr){1-7}
Model & Bias(\%) & UW & FQ & \#sat & \#unsat & Time (s)\\
\midrule
AC13 (Original) &9.33 &Y &180 &20 &6 &30490.02 \\
AC14 (ADF) &2.61 &Y &180 &12 &1 &24475.46 \\
AC15 (EIDIG-5) &1.00 &Y &180 &8 &1 &15673.36 \\
AC16 (EIDIG-$\infty$) &1.24 &Y &180 &12 &1 &22035.31 \\
\bottomrule
\end{tabular}

\vspace{0.9em}
\textbf{Artificially fair models; protected attribute: Age}\\[0.35em]
\begin{tabular}{@{}l c r r r r@{}}
\toprule
\multicolumn{6}{c}{PyFair}\\
\cmidrule(lr){1-6}
Model & Fair & FQ & \#sat & \#unsat & Time (s)\\
\midrule
fair\_dual\_GC1 &Unk &102 &86 &145 &1800\\
fair\_dual\_GC2 &Unk &202 &37 &175 &1800\\
fair\_dual\_GC3 &Y &20 &1835 &227 &1798\\
fair\_dual\_GC4 &Y &22 &382 &28 &40\\
fair\_dual\_GC5 &Unk &248 &1 &2 &1800\\
\bottomrule
\end{tabular}
\end{table}

  We test these four models using our fairness checking framework to evaluate how effectively PyFair identifies discriminatory instances in retrained models.
  Although ADF and EIDIG reduce the sampled Bias(\%) metric relative to the original model, PyFair still finds at least one counterfactual unfairness witness in each retrained model. In Table~\ref{Tab:pyct_EIDIG}, we present the results of using PyFair to analyze the original model and the retrained models. The results indicate that lower sampled Bias(\%) does not imply the absence of counterfactual unfairness witnesses.

\paragraph{RQ4: How effective is PyFair in verifying absence of witnesses under the encoding?}

  To answer this question, we evaluate PyFair's effectiveness on artificially fair models. These models are derived from the benchmarks by setting the outgoing edge weights to zero for input nodes on PAs, ensuring that these attribute values do not influence the model outcome. Also, the Sigmoid function in the output layer is replaced with a direct comparison with the threshold, allowing PyFair to faithfully encode the relevant decision predicate in linear real arithmetic.
  % Finally, in addition to setting the PA outgoing weight to zero, we modified the output layer activation function to use a weighted sum and employed zero as the threshold for classification. 
  %
  %The experimental results demonstrate that PyCT is effective in verifying the fairness of ReLU-based models.  under ReLU-based architectures

%\input{nccu-thesis/tables/fair_GC_model_PA_compare}

\begin{table}[t]
\centering
\caption{Summary of conditional verification outcomes in RQ4, derived from the artificially fair-model block of Table~\ref{Tab:pyct_EIDIG}.}
\label{tab:rq4-summary}
\begin{tabular}{lr}
\toprule
Outcome & Count \\
\midrule
Artificially fair models evaluated & 5 \\
Verified fair under the encoding & 2 \\
Timeout/inconclusive & 3 \\
\bottomrule
\end{tabular}
\end{table}

The experimental results reveal both the potential and limitations of our tool in verifying absence of witnesses, as only relatively small models can be verified fair under the encoding within the 1800-second timeout. %This limitation is evident in the AC, BM, and GC model sets, where larger models often result in an ``Unk'' (unknown) fairness status.
%For example, in the right table in Table~\ref{Tab:pyct_EIDIG}, the fair models verified by PyFair are relatively small (which can be seen by their small FQ values). For the GC5 model, PyFair only manages to explore one test case due to costly constraint solving.
Thus, even though PyFair can verify the absence of witnesses when the corollary's premises hold, the runtime may be considerable for complex models. This scalability issue underscores the need for more efficient techniques to handle the path constraints generated during fairness verification.

\subsection{Limitations}

The guarantees provided by PyFair are conditional on the encoded semantics. A verification result applies to the classifier decision predicate and input-domain constraints represented in SMT; it does not automatically account for floating-point implementation behavior unless such behavior is modeled. The validity of generated witnesses also depends on correctly encoding feature-domain constraints, especially for categorical, integer, and normalized attributes. In addition, concolic path exploration is subject to path explosion, so timeout-limited runs that do not exhaust the queue are inconclusive rather than fairness proofs. Finally, the artificially fair models used for sanity-check verification are useful for evaluating the verification workflow but should not be interpreted as representative of all deployed fair classifiers.

\FloatBarrier

\section{Conclusion}
\label{c:conclusion}
This work proposes PyFair, a framework for automatic protected-attribute counterfactual unfairness witness search in neural-network classifiers. PyFair synthesizes paired inputs through concolic path exploration and SMT solving, using a dual-network construction to search for pairs that share non-protected attributes but receive different classifier labels. When the classifier decision predicate and input-domain constraints are exactly encoded and concolic exploration is exhaustive, PyFair can verify the absence of such witnesses under the encoding.
As with other constraint-based testing approaches \cite{yu2024constraint,biswas2023fairify}, our evaluation reveals limitations around scalability. The experiments show strong witness-finding performance on existing benchmarks and highlight path explosion as the main obstacle to broader verification. Future work could enhance the framework through more efficient algorithms for handling larger architectures, extension to different network types like RNNs and Transformers, and deeper integration with fair model training techniques.

% \section*{ACKNOWLEDGMENT}
% This is the Acknowledgment section. Note that it is unnumbered.
\vspace{1em}

\footnotesize
\FloatBarrier
\bibliographystyle{JISEbib}
\bibliography{reference}

\end{document}